\documentclass[cameraready]{Interspeech}

\title{Turning Speech Language Models into Multilingual Listeners}

\author[affiliation={1}, orcid=0000-0000-0000-0000,correspondingauthor]{Tolulope}{Ogunremi}
\hypersetup{pdfauthor={Tolulope Ogunremi}}
\author[affiliation={1}, orcid=0000-0002-6459-7745]{Dan}{Jurafsky}
\author[affiliation={1}, orcid=0000-0001-6155-649X]{Christopher D.}{Manning}
\author[affiliation={2,3}, equalcontribution]{Ahnmet}{Üstün}
\author[affiliation={1}, orcid=0000-0003-1006-8669, equalcontribution]{Martijn}{Bartelds}

\address{
    $^1$ Department of Computer Science, Stanford University, United States \\
    $^2$ Cohere Labs, Canada \\
    $^3$ Cohere, Canada
}

\email{\{tolulope,bartelds\}@cs.stanford.edu, ahmet@cohere.com}


\usepackage{comment}

\definecolor{citecolor}{HTML}{2779af}
\definecolor{linkcolor}{HTML}{c0392b}
\definecolor{urlcolor}{HTML}{904080}
\usepackage{url}

\usepackage{booktabs}
\usepackage{amssymb}
\usepackage{url}
\usepackage{array}
\usepackage{float}
\usepackage{cite}

\usepackage{cancel}

 \usepackage{graphicx}
\usepackage{caption}
\usepackage{subcaption}
 \usepackage{IEEEtrantools}

\usepackage{hyperref}
\hypersetup{pdfauthor={Tolulope Ogunremi}}

\newcommand{\dataset}{\textsc{MultiSpeechQA}}
\newcommand{\eval}{\textsc{MultiSpeech-Bench}}
\newcommand{\model}{\textsc{MultiSpeech}}

\begin{document}
\bstctlcite{IEEEexample:BSTcontrol}

\maketitle
\begin{abstract}
Speech Language Models (SLMs) that understand spoken language questions support only a few high-resource languages, limiting access to millions of people worldwide. This gap stems from the scarcity of multilingual speech instruction-tuning datasets. We present MULTISPEECHQA, a large-scale, synthetically generated and human-verified dataset comprising 9200 hours of 10.8 million spoken question-answer pairs in 23 typologically diverse languages. Using MULTISPEECHQA, we also introduce MULTISPEECH-BENCH, a multi-task benchmark for evaluating SLM performance on 23 languages. We compare the performance of a cascading system to open-weight and closed SLMs on MULTISPEECH-BENCH and find that the cascading system outperforms open-weight SLMs but not all closed SLMs. We use MULTISPEECHQA to finetune Qwen 2.5-Omni, which improves its performance on our benchmark. Our findings show that high-quality synthetic datasets offer a cheap solution to improving the multilingual capabilities of SLMs.
\end{abstract}

\section{Introduction}

Speech Language Models (SLMs) often combine a pretrained speech encoder with a pretrained Large Language Model (LLM), using a modality adapter module to map the output of the speech encoder into the language model input space to perform various speech and language processing tasks~\cite{arora2025landscapespokenlanguagemodels}. These models are trained with instruction tuning data to align the speech encoder and LLM, and allow for natural spoken interactions. 

SLMs have many advantages over alternatives like the popular multitask speech model Whisper \cite{radford2023robust}, including allowing natural language instructions for speech tasks, doing question answering out-of-the-box and enabling zero-shot performance in a variety of traditional speech processing tasks, such as emotion recognition, audio captioning or audio-based storytelling.
However, the open-weight SLMs that exist today are primarily developed for English and a few other high-resource languages \cite{zhang2023speechgpt,chu2024qwen2,fang2024llama,salmonn,abouelenin2025phi}. This limits access to the state-of-the-art speech-language technology for many speakers worldwide.

The most critical challenge in developing multilingual SLMs is the scarcity of multilingual speech-language instruction-tuning datasets. While there has been significant progress on curating such multilingual data for text-only models \cite{singh-etal-2024-aya,ustun-etal-2024-aya}, and vision-language models \cite{dash2025aya,yue2025pangea}, the intersection of speech and language remains severely limited. 

There are several benchmarks that have been introduced to measure speech language model capabilities. These include speech and audio understanding in AudioBench~\cite{wang2024audiobench}, spoken language understanding with SLUE~\cite{shon2022slue},  long-form audio reasoning 
with BLAB~\cite{ahia2025blabbrutallylongaudio}, and safety, bias and fairness evaluation with AHELM~\cite{lee2025ahelm} and AIR-Bench~\cite{yang2024air}. Yet the existing evaluation benchmarks for SLMs suffer from a lack of language coverage, all only containing complex tasks in English, and this lack is particularly problematic for complex tasks like open-ended generative instruction following.

To address this gap, we present \dataset{}, a large-scale multilingual spoken question-answering (SQA) dataset comprising of 10.8 million instructions and 9200 hours of synthetically generated and human-verified speech data in 23 typologically diverse languages. \dataset{} consists of open-ended question-answer pairs from variety of tasks, and data sources designed to foster instruction following capabilities for SLMs in 23 languages. 
We combine \dataset{} with CommonVoice~\cite{commonvoice} automatic speech recognition (ASR) data and CoVoST-2~\cite{wang2021covost} automatic speech translation (AST) data to create \eval{}, providing a multi-task evaluation suite of these models in 23 languages.

Our main contributions are as follows:

\begin{enumerate}
\item We validate the hypothesis that automated synthetic data generation can provide sufficiently good instruction-tuning data to enable effective post-training of SLMs for many languages, provided only that adequate machine translation (MT) and speech synthesis (TTS) systems exist for those languages.
    \item We provide \dataset{}, a \textbf{multi}lingual speech-language instruction fine-tuning dataset that consists of over 10.8 million \textbf{spoken question-answer} pairs in 23 languages, where multilingual samples are generated by using translation and speech synthesis, comprising 9200 hours in total.   
    \item We develop \eval{}, a \textbf{multi}lingual, \textbf{multi}task \textbf{speech} processing benchmark, facilitating evaluation of speech recognition, speech translation and spoken question answering in 23 languages.
    \item Validating the effectiveness of our dataset, we finetune Qwen2.5-Omni on \dataset{} and show that it achieves state-of-the-art performance among open-weight models on \eval{}, particularly outperforming Qwen2.5-Omni with 60\% win-rate across 23 languages.
\end{enumerate}

By releasing our dataset and model weights, we aim to extend the benefits of modern speech technology to speakers of diverse languages worldwide. Our dataset, benchmark and models are publicly available.
\footnote{\url{https://multispeech.github.io}}

\section{Related Work}
\noindent\textbf{Speech Language Models.} SLMs can be broadly categorized into three architectural approaches: (1) models of speech distribution; (2) models of joint speech-text distribution, and (3) models combining pre-trained text LLMs with speech encoders~\cite{arora2025landscapespokenlanguagemodels}.
The third approach leverages the instruction-following capabilities learned by the text LLM and typically requires less training data, enabling strong few-shot or zero-shot performance on a variety of multimodal tasks~\cite{chen2024voicebenchbenchmarkingllmbasedvoice}.
Many state-of-the-art models adopt this approach, including proprietary models, such as Gemini 2.5~\cite{comanici2025gemini} and GPT-4o~\cite{openai2024gpt4technicalreport}, as well as notable open-weight models, such as Phi-4-Multimodal~\cite{abouelenin2025phi}, SALMONN~\cite{salmonn}, and Qwen2Audio~\cite{chu2024qwen2}. and Qwen 2.5 Omni~\cite{Qwen2.5-Omni}. 

Comparing open-source models reveals limited multilingual support. SALMONN is primarily trained on English data, while Phi-4-Multimodal, Qwen2Audio and Qwen 2.5 Omni support only eight languages. SALMONN, Qwen2Audio and Qwen 2.5 Omni leverage a Whisper-based encoder~\cite{radford2023robust}, which is aligned with an LLM backbone, suggesting potential for broader language coverage that remains largely unexplored. Our work substantially extends the language coverage of these models by providing support for 23 languages with a comprehensive evaluation. \\

\noindent\textbf{Multilingual SQA Datasets.} Multilingual SQA datasets are scarce, limiting the development of truly multilingual SLMs. Existing multilingual speech benchmarks and datasets primarily target traditional tasks, rather than open-ended SQA. For example, ASR and AST dataset FLEURS~\cite{fleurs} and the ASR dataset ML-SUPERB~2.0~\cite{shi24g_interspeech} cover 102 and 143 languages, respectively. For SQA specifically, Voice Assistant 400K~\cite{xie2024mini} offers diverse question-answer pairs but only in English. 

We draw on recent work showing that high-quality synthetic speech can effectively augment limited real data, such as Phi-4-Multimodal demonstrated strong performance in SQA tasks using synthetic speech from translations. Our training approach leverages this insight, while substantially expanding the language coverage with \dataset{}.

\section{Dataset Creation}
\begin{figure*}[t]
    \centering
    \includegraphics[width=0.9\linewidth]{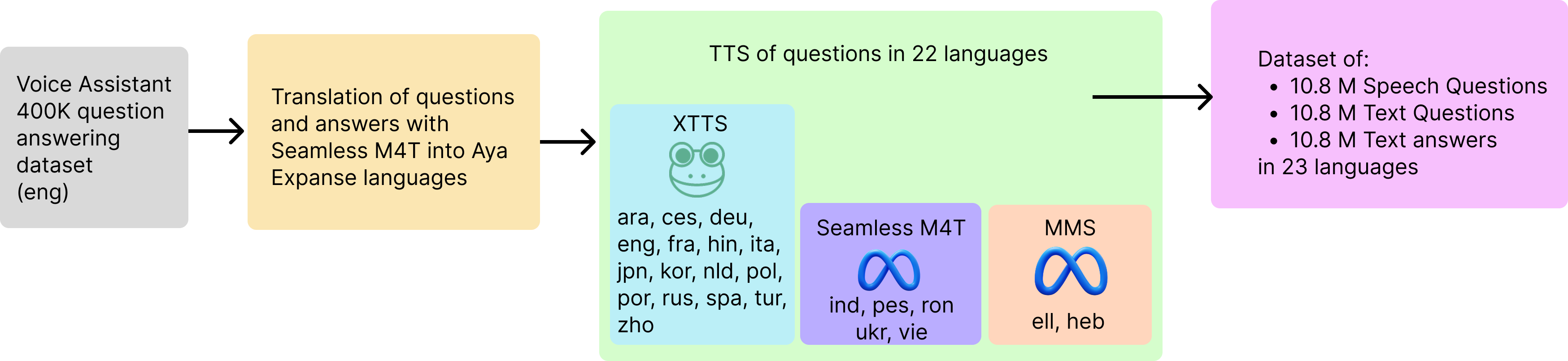}
    \caption{Summary of the dataset creation process. We translate the questions and answers of the Voice Assistant 400K dataset into each of the Aya Expanse languages (in Table \ref{tab:languages_combined}) with SeamlessM4T, synthesise the questions with XTTS for the languages covered by the model and SeamlessM4T for the languages not covered by XTTS, leaving us with roughly 10.8 million text questions, speech questions and text answers.}
    \label{fig:overview}
\end{figure*}

Figure~\ref{fig:overview} presents our two-stage process to create \dataset{}. We build upon the English Voice Assistant 400K (VA 400K; \cite{xie2024mini}) dataset, which consists of synthesised speech from text-only instruction-completion pairs.
These instruction-completion pairs are sourced from multiple datasets, as detailed in Table~\ref{tab:dataset_split}. These include the Alpaca GPT-4 datasets, instruction-tuning data generated with Alpaca questions with GPT-4, Trivia datasets adapted for question answering (Trivia single choice and Trivia multi choice), question answering datasets created for dialogue systems (QA assistant V1 and V2), questions and answers regarding a voice assistants identity (Identity) and Anthropic's Helpful Harmful dataset (RLHF).

We extend VA 400K to 22 additional languages covered by the Aya Expanse 8B~\cite{dang2024ayaexpansecombiningresearch} model through translation and synthesis, which has been shown to be effective in past work~\cite{abouelenin2025phi}. We summarise the languages in our dataset in Table \ref{tab:languages_combined}.

\begin{table}[ht]
\centering
\caption{Splits in our dataset drawn from the 470k entried in the VA 400K dataset translated into 22 languages with number of instruction-completion pairs per language.}
\begin{tabular}{lr}
\toprule
\textbf{Dataset} & \textbf{Number of pairs per language} \\
\midrule
Trivia (Multi-choice, 17K) \cite{trvia-single-choice} & 16,528 \\
Trivia (Single-choice, 20K) \cite{trvia-single-choice} & 16,529 \\
QA Assistant V1 (7K) \cite{qa-assitant}& 5,769 \\
QA Assistant V2 (20K) \cite{qa-assitant-2} & 16,008 \\
Alpaca GPT-4 (EN, 55K) \cite{peng2023instruction} & 31,293 \\
Identity \cite{xie2024mini} & 4,306 \\
RLHF \cite{hh-rlhf-anthropic}{} & 379,621 \\
\midrule
\textbf{Total} & \textbf{470,054} \\
\bottomrule
\end{tabular}
\label{tab:dataset_split}
\end{table}

\begin{table*}[h!]
\centering
\scriptsize
\setlength{\tabcolsep}{5pt}
\caption{Languages in \dataset{} with their language families, ISO 639-1 codes, TTS model used, and human evaluation scores for naturalness and content understanding.}
\begin{tabular}{llllcc}
\toprule
\textbf{Language} & \textbf{Language Family} & \textbf{ISO 639-1} & \textbf{TTS Model} & \textbf{Naturalness} & \textbf{Content Understood} \\
\midrule
Arabic & Afro-Asiatic (Semitic) & ara & XTTS & 2.8 & 3.7 \\
Chinese (Simplified) & Sino-Tibetan (Sinitic) & zho & XTTS & 2.8 & 4.8 \\
Czech & Indo-European (Slavic, West) & ces & XTTS & -- & -- \\
Dutch & Indo-European (Germanic, West) & nld & XTTS & 3.1 & 4.4 \\
English & Indo-European (Germanic, West) & eng & -- & -- & -- \\
French & Indo-European (Romance) & fra & XTTS & 3.6 & 4.3 \\
German & Indo-European (Germanic, West) & deu & XTTS & 3.0 & 4.4 \\
Greek & Indo-European (Hellenic) & ell & MMS & 2.4 & 4.2 \\
Hebrew & Afro-Asiatic (Semitic) & heb & MMS & 2.1 & 2.4 \\
Hindi & Indo-European (Indo-Aryan) & hin & XTTS & 3.4 & 3.5 \\
Indonesian & Austronesian (Malayo-Polynesian) & ind & XTTS & 3.0 & 4.1 \\
Italian & Indo-European (Romance) & ita & XTTS & 3.5 & 4.5 \\
Japanese & Japonic & jpn & XTTS & 3.0 & 2.9 \\
Korean & Koreanic & kor & XTTS & 2.3 & 4.2 \\
Farsi & Indo-European (Iranian) & pes & Seamless & 2.5 & 4.2 \\
Polish & Indo-European (Slavic, West) & pol & XTTS & 4.0 & 4.4 \\
Portuguese & Indo-European (Romance) & por & XTTS & 3.7 & 4.6 \\
Romanian & Indo-European (Romance) & ron & Seamless & 1.8 & 4.0 \\
Russian & Indo-European (Slavic, East) & rus & XTTS & 3.7 & 4.6 \\
Spanish & Indo-European (Romance) & spa & XTTS & 3.6 & 4.9 \\
Turkish & Turkic (Oghuz) & tur & XTTS & 3.4 & 4.3 \\
Ukrainian & Indo-European (Slavic, East) & ukr & Seamless & 2.5 & 4.6 \\
Vietnamese & Austroasiatic (Vietic) & vie & Seamless & 3.2 & 2.8 \\
\bottomrule
\end{tabular}
\label{tab:languages_combined}
\end{table*}

\subsection{Translation and Synthesis}
For translation, we use Seamless M4T v2 Large \cite{seamless2023} to translate instruction-completion pairs from English into the 22 target languages languages. This model was chosen as it is publicly available, free to use, and it achieves stronger performance compared to other models of similar size, such as NLLB \cite{team2022no}, in our preliminary experiments.

For speech synthesis, we use different models based on language support. We use XTTS \cite{casanova2024xtts} for 15 languages, as we found its audio quality to be superior to other models in our preliminary evaluations.
For the remaining seven languages not supported by XTTS, we use Seamless M4T v2 Large and language-specific MMS text-to-speech (TTS) models \cite{pratap2024scaling}.
To improve speaker diversity in the training data, which is important for achieving robust performance (e.g., see~\cite{NEURIPS2018_6832a7b2}), we leverage XTTS's voice cloning capability with short LibriVox \cite{McGuire_2005} clips of perceived male and female speakers.
For each language supported by XTTS, we randomly select a voice, which might be male or female, from all the LibriVox clips during synthesis, resulting in 37 different voices across the dataset. Table \ref{tab:languages_combined} shows the model assignment per language.

\subsection{Human evaluation}

To measure both the quality of the translations and synthesised speech, we conduct a human evaluation for our dataset. We sample 20 instruction-completion pairs for each language from our dataset, and ask native speakers of each language to evaluate both the naturalness of the speech and the amount of content they have understood on a 5-point scale (1 being least natural or least understood and 5 being extremely natural and understandable. We ensure that each language's examples were reviewed by at least two native speakers, except for Czech for which we could not obtain any ratings.

As shown in Table \ref{tab:languages_combined}, scores for the perceived naturalness of the speech range from 1.8 to 4.0, and the scores for content understanding range from 2.4 to 4.9. Unsurprisingly, the average score for naturalness (3.0) falls behind the content understanding (4.1), as the speech synthesis models often struggle to generate the highest quality natural sounds in many languages~\cite{casanova2024xtts, pratap2024scaling}.

Comparing the models used for speech synthesis, XTTS shows better performance than SeamlessM4T and language-specific MMS TTS models, achieving an averaged score of 3.3 and 4.2 in 15 languages for naturalness and the amount of content understood, respectively. Results for the language-specific MMS TTS models are 2.25 and 3.3 averaged across two languages, and SeamlessM4T are 2.5 and 3.9. Note that the languages that use MMS TTS models where XTTS does not have language coverage are lower-resource languages such as Farsi and Greek. These results show that our dataset is adequate for multilingual instruction finetuning, while further improvements will most strongly depend on improving TTS quality.

\subsection{\eval{} for Multilingual and Multitask Evaluation}
We split the \dataset{} dataset into train, development and test sets. We randomly sample SQA pairs of the different subsets with the same dataset distribution as VA 400K, resulting in a development set of 2000 SQA pairs and a test set of 1000 SQA pairs. To avoid speaker overlap, we select different speakers for the train and test set when finding new voice prompts is possible. All remaining data belongs to the train set.

We create \eval{} from a subset of our test split, sampling the same 200 SQA pairs per language. To ensure the quality of this evaluation dataset, we collect human annotations on all 200 SQA pairs using Prolific, asking language experts to review and correct the translations where necessary. Overall, 72\% of translations required editing with language-specific correction rates ranging from 43\% (Turkish) to 86\% (Chinese). This manually verified subset is combined with existing test from CommonVoice (ASR) and CoVoST-2 (ASR) for matching languages, creating our multilingual, multitask benchmark.

\begin{figure*}
    \centering
    \includegraphics[width=0.9\linewidth]{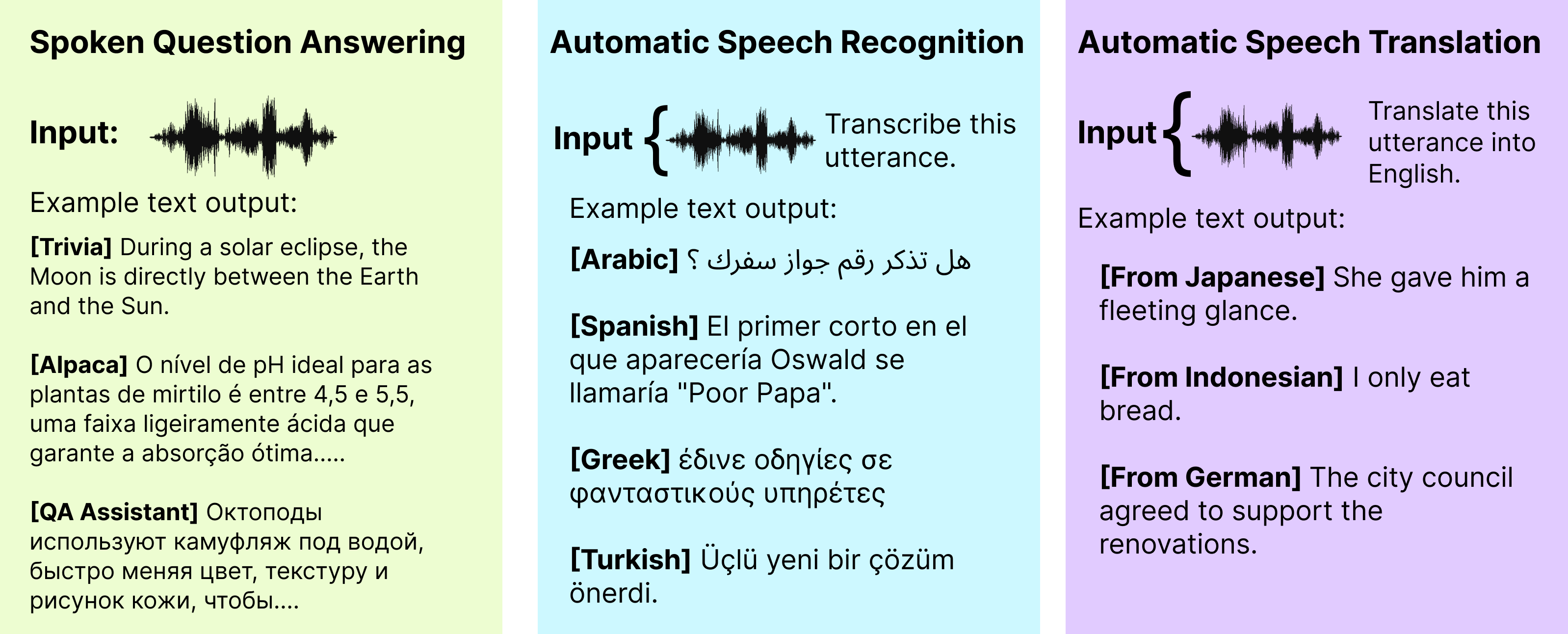}
    \caption{\eval{} covers three tasks: (1) Spoken Question Answering (SQA), where models are prompted with speech only (no text prompt); (2) Automatic Speech Translation (AST) from speech to English, using CoVoST-2 (X to En)  for languages that overlap with the 23 languages in our dataset; and (3) Automatic Speech Recognition (ASR) on CommonVoice for each of the 23 supported languages.}
    \label{fig:tasks}
\end{figure*}

\section{Evaluation on Open-Weight and Commercial Models}
To establish the performance of current multilingual SLMs, we evaluate leading open-weight SLMs on \eval{} and compare it against a strong cascading system baseline.
This evaluation quantifies the performance gap between languages and establishes baselines for measuring SQA performance improvements from training with \dataset{}.
For the SQA portion of \eval{}, we adopt pairwise preference evaluations using LLM-as-a-judge, following recent work involving open-ended multilingual generation~\cite{ustun-etal-2024-aya}. This approach allows for consistent evaluation across our 23 languages, and is more cost-efficient than recruiting human annotators for each language.
We use the multilingual Command-A \cite{cohere2025command} model as our LLM-as-a-judge, which supports the 23 languages in our datasets. To check for calibration across LLMs, we also use GPT-4o as an LLM judge.

\subsection{Cascading System Baseline:} While end-to-end models that process speech directly have architectural advantages (e.g., preserving acoustic information), we include a strong cascading system baseline by first transcribing the speech with Whisper Large v3 \cite{radford2023robust}, then prompting Aya Expanse 8B \cite{dang2024ayaexpansecombiningresearch} with the transcription.

Whisper is a leading multilingual model that supports over 96 languages and is trained to do ASR and AST into English. Aya Expanse 8B is a language model trained to respond to questions in all 23 languages in \dataset{}.
Such cascading baselines perform often on par or exceed the performance of SLMs on some spoken language processing tasks~\cite{chen2024voicebenchbenchmarkingllmbasedvoice}. Our benchmark can help us learn whether this is true for our three tasks, even though non-cascading SLMs have many other advantages -- for example, they are the only choice for performing speech-native tasks like spoken emotion detection or speaker identification.

\subsection{Open-Weight Models:} We evaluate three leading open-weight SLMs that represent different training approaches and cover different languages:
\begin{enumerate}
    \item \textbf{Qwen2-Audio} \cite{Qwen2-Audio}: A speech-aware language model that combines an audio encoder initialised from Whisper Large v3 \cite{radford2023robust} with a QwenLM 7B decoder \cite{chu2024qwen2}. The modality adapter is a multi-layer perceptron. The authors do not specify full language support, but model performance is reported on English, French, and Chinese.
    \item \textbf{Qwen2.5-Omni} \cite{Qwen2.5-Omni}: A multimodal model incorporating speech and vision modalities into the Qwen2.5 language model. The processing of multimodal inputs and text generation happens in the `Thinker' part of the model. For speech, it uses an encoder that is initialized with Whisper Large v3, and a multi-layer perceptron as the modality adapter. The languages supported by the model are not explicitly stated. 
    \item \textbf{Phi-4-Multimodal} \cite{abouelenin2025phi}: A multimodal model incorporating speech and vision modalities into the Phi-4 language model. For speech, it uses a conformer model, which is trained on a proprietary dataset. The modality adapter is a multi-layer perceptron. The model supports English, Chinese, German, French, Italian, Japanese, Spanish, and Portuguese audio input.
\end{enumerate}

\subsection{Commercial SLMs:} We evaluate four commercial SLMs: \textbf{GPT-Audio Gemini 2.5 Flash, Flash Lite and Flash Pro}. GPT-Audio is OpenAI's speech-enabled variant of the GPT family, designed for real-time multimodal interaction. It supports speech recognition, speech-to-text reasoning, and text-to-speech generation within a unified model. The Gemini 2.5 Series of models provide robust automatic speech recognition and basic audio-event understanding.

\subsection{Results}
\begin{figure*}
    \centering
    \includegraphics[width=0.9\linewidth]{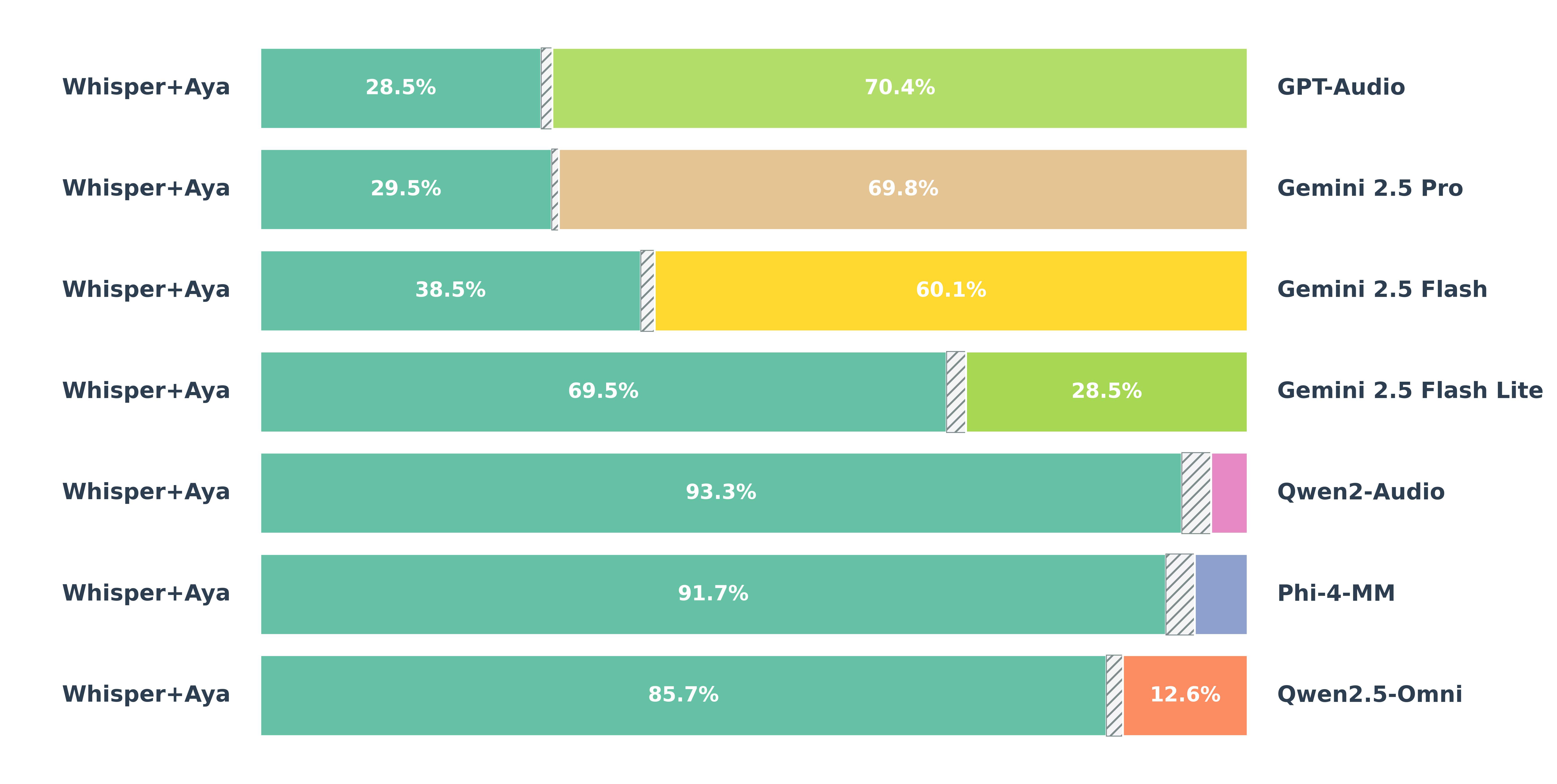}
    \caption{Win rates on \eval{} averaged across languages for the open-weight and commercial SLMs against the baseline cascading system of Whisper combined with Aya Expanse 8B using the Command-A LLM-as-a-Judge. Bars show \% wins for each model and \% ties (grey lines).}

\label{fig:slms-vs-whisper-aya-command-a}
\end{figure*}

\begin{figure*}
    \centering
    \includegraphics[width=0.9\linewidth]{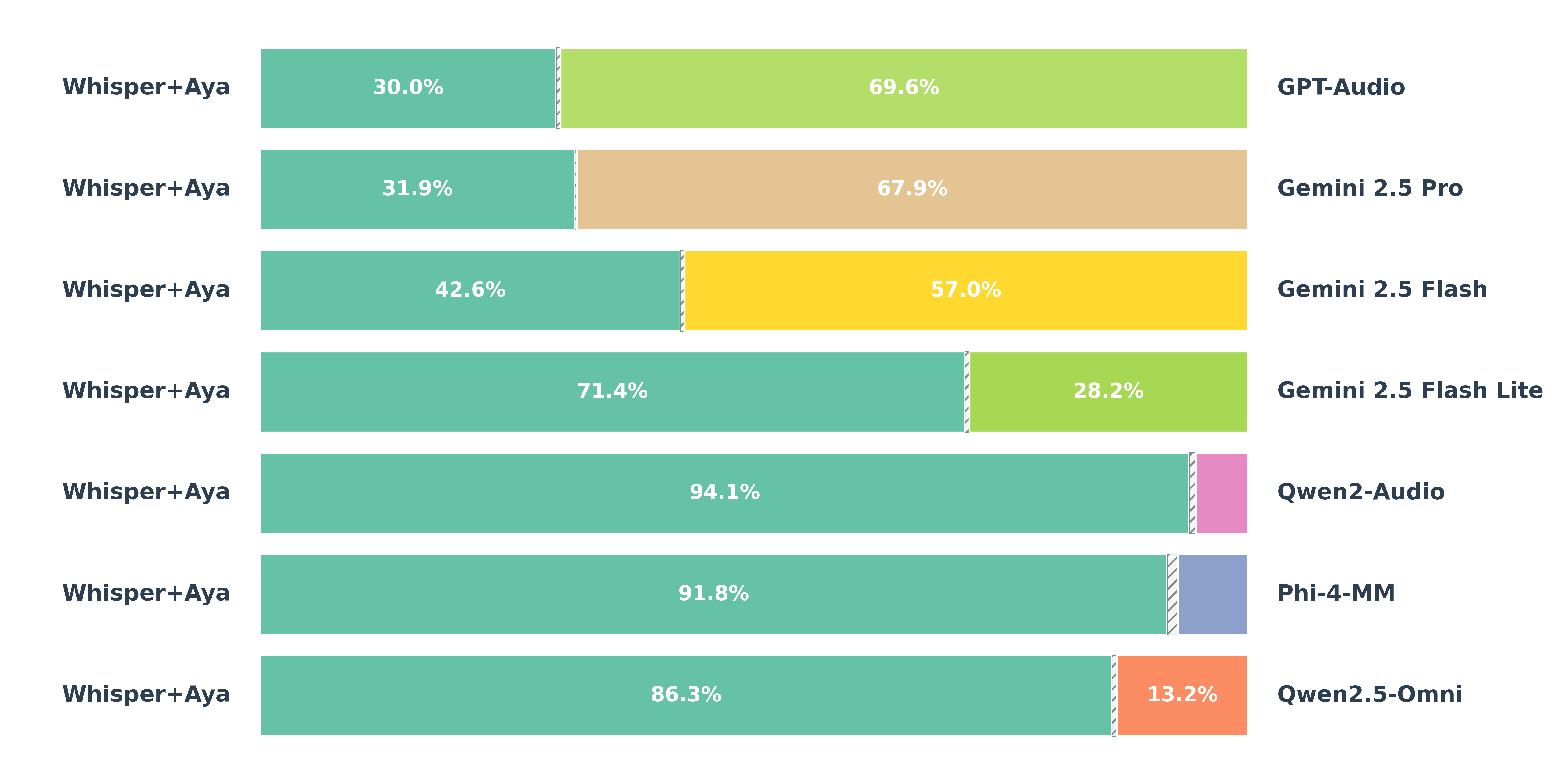}
    \caption{Win rates on \eval{} averaged across languages for the open-weight and commercial SLMs against the baseline cascading system of Whisper combined with Aya Expanse 8B using the GPT-4o LLM-as-a-Judge. Bars show \% wins for each model and \% ties (grey lines).}

\label{fig:slms-vs-whisper-aya-gpt-4}
\end{figure*}

With Whisper combined with Aya Expanse 8B as the baseline, Figure~\ref{fig:slms-vs-whisper-aya-command-a} shows win rates on the SQA portion of \eval{} for each open-weight and commercial models tested against it with Command-A as an LLM judge. Figure~\ref{fig:slms-vs-whisper-aya-gpt-4} shows win rates with GPT-4o as an LLM judge.
We find that Qwen2.5-Omni outperforms all other open-weight models, and our analysis of language-specific win rates reveals that SLMs perform better on the languages they explicitly support.
Qwen2-Audio and Phi-4-Multimodal are competitive with the baseline in languages that the models are trained on, but it is clear that they are outperformed by the cascading system baseline, likely due to the strength of the individual ASR and language models on their specific tasks and the lack of catastrophic forgetting that can occur during instruction tuning of the models to enable multimodal processing. Almost all closed-weight models beat the Whisper+Aya baseline, with the exception of Gemini-2.5-Flash Lite. GPT-Audio is the best performing commercial SLM with Gemini 2.5 Pro closely following.

In Table~\ref{tab:asr_ast_results_all}, we show the performance of the baseline and open-weight SLM models on ASR and AST, measuring ASR performance using the word error rate (WER; character error rate (CER) for Chinese (zh) and Japanese (ja)) and AST performance with BLEU~\cite{papineni-etal-2002-bleu} and chrF~\cite{popovic-2015-chrf}. Qwen2.5-Omni shows the strongest performance among the evaluated SLM models (average ASR error rate of 49.7; average BLEU of 22.7; average chrF of 46.6), outperforming the baseline on AST.
The per-language results are mixed for both tasks, but generally models perform strongest on the languages seen during training.

\begin{table*}[t]
\caption{Speech Recognition (ASR) and Speech Translation (AST) performance across models and languages. The baseline results are for the cascaded Whisper and Aya-Expanse 8B model. We compare the baseline to Qwen2-Audio (Q2-Audio), Phi-4-Multimodal (Phi-4 MM), and Qwen2.5-Omni (Q2.5O). For AST, we select the languages in the dataset that have both X $\rightarrow$ En and En $\rightarrow$ X directions. We report CER for Chinese (zh) and Japanese (ja), and report chrF in addition to BLEU for AST performance.}
\centering
\scriptsize
\addtolength{\tabcolsep}{-3pt}
\begin{tabular}{l | ccccc | ccccc}
\toprule
\textbf{Lang.} 
& \multicolumn{5}{c|}{\textbf{ASR (WER \%) $\downarrow$}} 
& \multicolumn{5}{c}{\textbf{AST (BLEU / chrF) $\uparrow$}} \\
\cmidrule(lr){2-6} \cmidrule(lr){7-11}
& \textbf{Base} & \textbf{Q2-A} & \textbf{Phi4} & \textbf{Q2.5O} & \textbf{Q2.5O FT}
& \textbf{Base} & \textbf{Q2-A} & \textbf{Phi4} & \textbf{Q2.5O} & \textbf{Q2.5O FT} \\
\midrule
ar & 14.0 & 118.1 & 146.1 & 45.7 & 31.5 & 34.2/54.7 & 7.3/30.9 & 0.1/11.9 & 30.6/53.8 & 29.7/52.3 \\
cs & 26.0 & 117.4 & 115.2 & 100.3 & 101.6 & -- & -- & -- & -- & -- \\
de & 9.4 & 33.3 & 7.0 & 7.1 & 7.5 & -- & -- & -- & -- & -- \\
el & 21.8 & 118.3 & 114.1 & 108.0 & 108.3 & -- & -- & -- & -- & -- \\
en & 3.2 & 34.6 & 13.9 & 13.8 & 16.6 & -- & -- & -- & -- & -- \\
es & 7.7 & 18.1 & 4.9 & 4.9 & 5.1 & -- & -- & -- & -- & -- \\
fa & 38.0 & 128.7 & 133.0 & 109.1 & 107.5 & -- & -- & -- & -- & -- \\
fr & 9.0 & 34.1 & 10.1 & 10.9 & 10.8 & -- & -- & -- & -- & -- \\
he & 40.6 & 128.4 & 447.8 & 122.9 & 113.6 & -- & -- & -- & -- & -- \\
hi & 30.8 & 123.2 & 105.3 & 68.9 & 68.4 & -- & -- & -- & -- & -- \\
id & 34.9 & 71.8 & 125.1 & 14.0 & 13.9 & 36.1/53.3 & 6.6/29.8 & 0.2/15.0 & 37.0/59.0 & 34.7/57.9 \\
it & 5.0 & 21.7 & 5.1 & 6.5 & 6.2 & -- & -- & -- & -- & -- \\
ja & 15.8 & 66.1 & 78.6 & 75.9 & 31.3 & 10.4/23.4 & 11.3/38.5 & 19.7/46.6 & 17.8/41.5 & 17.2/41.2 \\
ko & 20.9 & 61.4 & 144.4 & 23.0 & 24.4 & -- & -- & -- & -- & -- \\
nl & 9.5 & 90.8 & 101.5 & 14.0 & 14.3 & -- & -- & -- & -- & -- \\
pl & 7.5 & 110.8 & 118.6 & 94.7 & 75.1 & -- & -- & -- & -- & -- \\
pt & 6.7 & 28.5 & 7.4 & 9.6 & 11.9 & -- & -- & -- & -- & -- \\
ro & 15.1 & 114.6 & 106.1 & 89.3 & 74.6 & -- & -- & -- & -- & -- \\
ru & 17.1 & 57.4 & 123.9 & 9.3 & 13.7 & -- & -- & -- & -- & -- \\
tr & 11.4 & 114.6 & 131.9 & 74.5 & 66.4 & 20.0/40.8 & 0.6/19.3 & 0.1/15.4 & 5.5/27.0 & 5.8/28.3 \\
uk & 18.7 & 107.0 & 118.8 & 82.6 & 81.3 & -- & -- & -- & -- & -- \\
vi & 18.0 & 110.5 & 104.2 & 50.9 & 169.3 & -- & -- & -- & -- & -- \\
zh & 28.9 & 93.9 & 7.9 & 6.2 & 6.8 & 4.5/16.9 & 15.6/45.8 & 8.9/39.7 & 22.7/51.2 & 17.7/46.1 \\
\midrule
\textbf{Ave.} & 17.8 & 82.8 & 98.7 & 49.7 & 50.4 & 21.0/37.8 & 8.3/32.9 & 5.8/25.7 & 22.7/46.6 & 21.0/45.2 \\
\bottomrule
\end{tabular}
\label{tab:asr_ast_results_all}
\end{table*}

\subsection{Human validation of LLM-as-a-Judge}
\label{sec:appendix-humam-llm-calibration}

In addition to checking for model calibration by using two models as an LLM judge, we selected eight typologically diverse languages (Arabic, German, Hebrew, Hindi, Korean, Portuguese, Turkish and Chinese) to measure whether human judgements on SLM vs baseline pairs align with Command-A’s judgements. Eight of 23 languages were chosen due to budget constraints. We evaluated an open model (Qwen2.5-Omni) and a commercial SLM (GPT-Audio) against the baseline and asked native speakers to judge the responses. Ensuring that we had three annotations for each pair of answers in the benchmark, we derived a consensus label from the three annotations and measured human–LLM alignment, observing 75.6\% agreement for Qwen2.5-Omni ($\kappa$= 0.186) and 52.4\% for GPT-Audio ($\kappa$ = 0.185). For GPT-Audio, annotators showed high disagreement as the outputs are of similar quality. Overall, these experiments provide evidence that our LLM-as-a-judge setup captures human preferences to a reasonable extent, especially for Qwen2.5-Omni. \footnote{The low value of $\kappa$ is an artifact of a high baseline probability of chance agreement, likely driven by a severe quality imbalance between the competing models, skewed towards the baseline when compared to Qwen 2.5 Omni and skewed towards GPT-Audio when compared to the baseline.}

\section{Finetuning With \dataset{}}
\label{sec:ft}

The open-weight model results show the need for further model improvement. We take the best-performing open-weight SLM, Qwen2.5-Omni and finetune it on \dataset{}. We perform LoRA finetuning~\cite{hu2022lora} on all linear modules in each transformer layer, using a rank of 32. We train for a fixed number of steps, equalling roughly 3 epochs of the data. We then evaluate its performance on \eval{}.

\subsection{Results}
\begin{figure}[ht!]
    \centering
    \includegraphics[width=\linewidth]{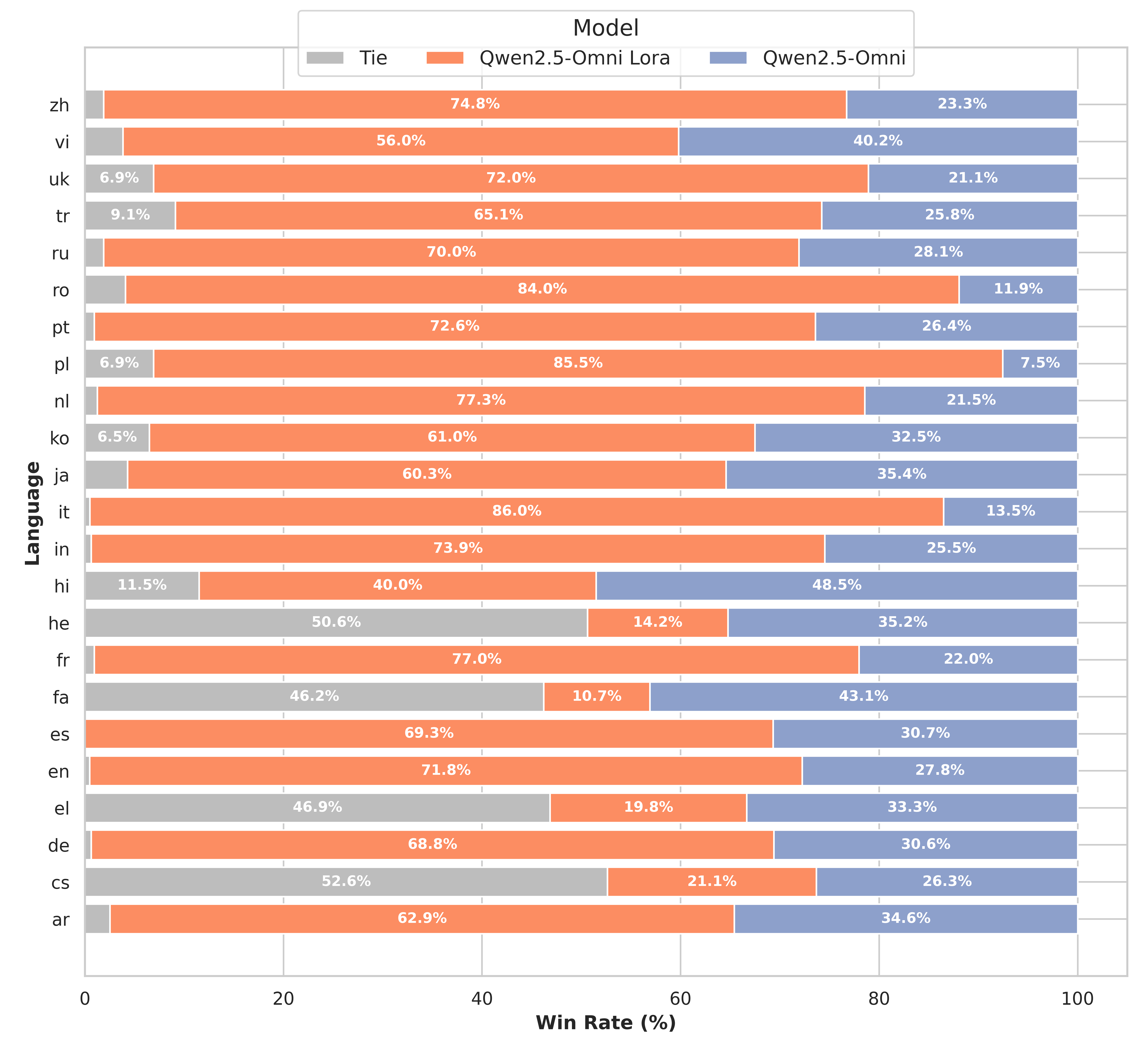}
    \caption{Win rates comparison: Qwen2.5-Omni vs. Qwen2.5-Omni finetuned. Finetuning improves performance across most languages, meaning that our \dataset{} enables better SQA capability. Hebrew, and Czech seem to lag behind, with most of the results being a tie.}
    \label{fig:winrate_lang_25lora_vs_25}
\end{figure}

\begin{figure}
    \centering
    \includegraphics[width=\linewidth]{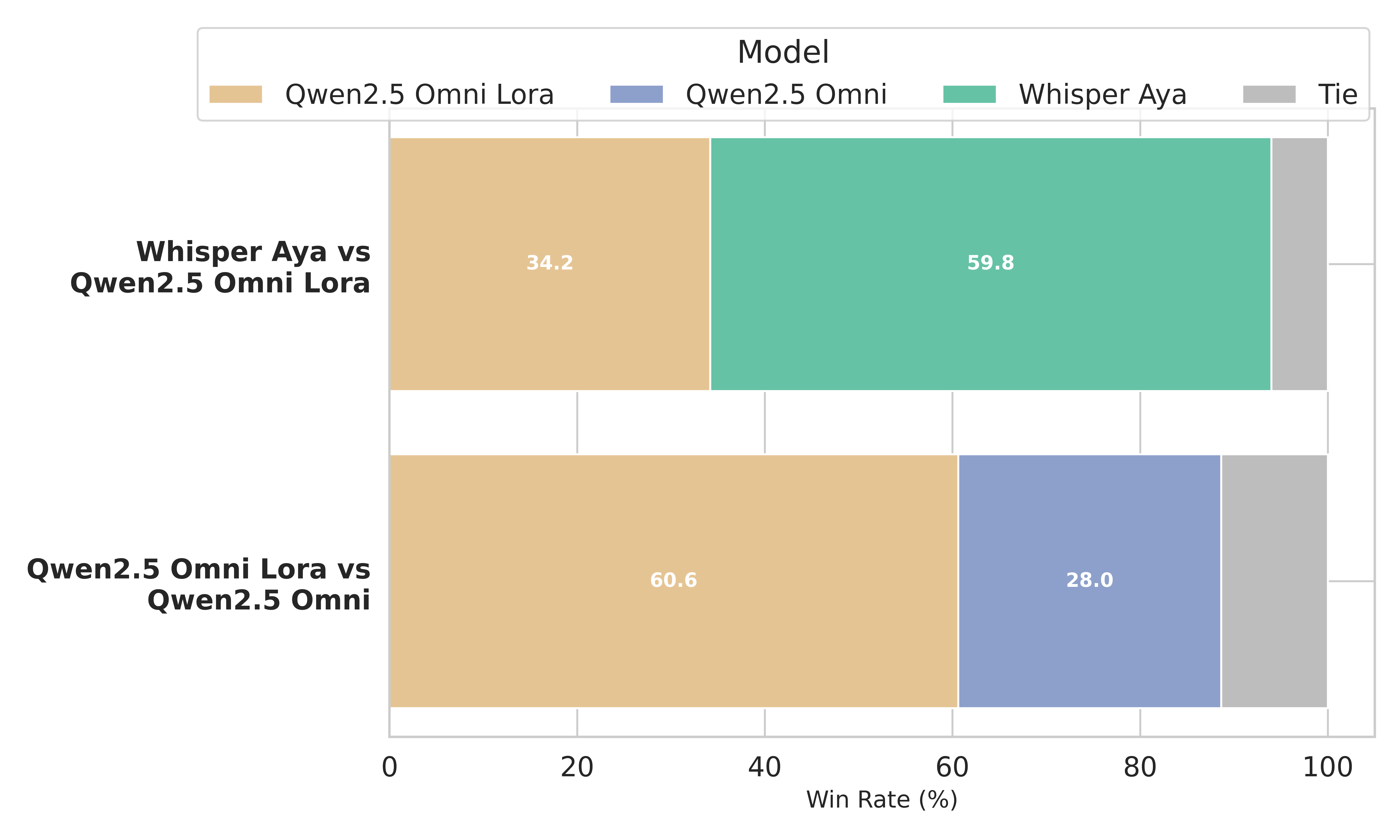}
    \caption{Win rates on \eval{} averaged across languages for the Qwen2.5-Omni and the Qwen2.5-Omni model finetuned on \dataset{}. Bars show \% wins for each model and \% ties (gray).
    }
    \label{fig:lora-ft-fig}
\end{figure}

Figure \ref{fig:lora-ft-fig} shows the win rates of Qwen2.5-Omni finetuned with \dataset{} against the non-finetuned Qwen2.5-Omni model and Figure~\ref{fig:winrate_lang_25lora_vs_25} shows the language breakdown of these results. We find that parameter efficient finetuning improves SQA performance substantially. The finetuned model wins the majority of the time, struggling with languages such as Hebrew, Greek and Farsi, where the judgements tie 48.0\% of the time on average. When considering all languages, our finetuned model wins 60.6\% of the time on average.
This finetuned model also performs best on SQA against the cascading baseline, leading to best performance among open-weight SLMs on the SQA portion of \eval{}.

Comparing the ASR and AST performance of Qwen2.5-Omni and Qwen2.5-Omni finetuned on \dataset{}, we find that the average ASR performance remains stable across languages (per-language results shown in Table~\ref{tab:asr_ast_results_all}). The average WER increases marginally from 49.7 for the non-finetuned model to 50.4 for the finetuned model.
On AST, the performance of the finetuned model is similarly comparable, as shown by the slightly lower BLEU score of 21.0 compared to 22.7 for the non-finetuned model. Overall, we find that \dataset{} finetuning substantially improves spoken SQA, while leaving performance on core ASR and AST capabilities effectively unchanged.

\section{How do training data mixtures affect speech language model performance?}
In Section~\ref{sec:ft}, we show that \dataset{} improves the SQA performance of Qwen2.5-Omni.
These results motivate a controlled study of data composition for multilingual SLMs.
Specifically, most existing SLMs, including Qwen2.5-Omni, are trained on undisclosed data mixtures, making it impossible to understand whether performance differences across languages arise from the model capacity being spread across many languages or from insufficient task diversity.
We therefore ask two questions: (1) Does training with fewer languages lead to better performance?; and (2) Does adding AST data improve model performance?

To answer these questions, we train models from scratch using the SALMONN \cite{salmonn} architecture, whose training code is publicly available. Specifically, in our setup, we use Whisper as the speech encoder and Aya Expanse 8B as the language model. The window-level Q-Former uses an mBERT text encoder. We choose this setup, because the Whisper encoder produces stable multilingual speech features, and the window-level Q-Former allows us to leverage a pretrained text encoder to more efficiently learn intermediary representations.

As Whisper is trained to translate speech data of many of the Aya's languages into English, we hypothesise that adding AST data increases the task diversity in the training mixture and hence could lead to a better downstream performance. For this ablation, we set a threshold of 20\% for the speech translation data and use mixed batches for more robust multi-task instruction-tuning.

\subsection{\model{} Models Training Details}
\label{sec:appendix-model-training-details}

\subsubsection{Stage 1 ASR training:} Following the SALMONN training setup, we train the window-level Q-Former and LoRA adapters using ASR data. 
To do this, we use a uniform amount of ASR data (20 hours) in all languages. In this stage, we add the text instruction ``Transcribe this utterance'' to the speech prompt. The goal of this stage is alignment between speech and text representations and enabling the model to understand the speech inputs.

We start with CommonVoice data \cite{commonvoice} for each language. Some of the languages have fewer than 20 hours of training data, so we balance the number of hours of data in Vietnamese with 15 hours of the Bud500 dataset \cite{Bud500}, 18 hours of Hebrew 
with the Ivrit.ai dataset \cite{marmor2023ivritai}, 14 hours Hindi with the monolingual portions of the  Multilingual and Code-Switching ASR Challenges for Low Resource Indian Languages dataset \cite{diwan2021multilingual} and 19 hours Korean with the Zeroth-Korean corpus \cite{Jo2022ZerothKorean}. 

\subsubsection{Stage 2 Question Answering Training:}
For the second stage of training, we use our \dataset{} dataset for all 23 languages. For the Trivia QA, the QA Assistant, and the Alpaca GPT-4 datasets, we use all the samples, but given the difference in distribution of Anthropic-RLHF data, we only subsample 1000 examples from this data source to ensure a training mixture that is balanced and optimized for general-purpose speech instruction-following tasks. Overall, our training mixture includes 2,070,000 samples distributed equally between 23 languages. 

In addition to \dataset{}, we also run ablations where we include additional speech translation (AST) data from the CoVoST-2~\cite{wang2021covost} dataset to the training mixture.

Finally, although the model architecture allows us to append a text prompt to the speech input, we train the model without any additional text prompt to enable the question answering capability from the spoken questions alone. 
\begin{table}[ht!]
    \centering
    \vspace{0.2cm}
    \caption{Hyperparameters used to train Stage 1 and Stage 2 of \model{} models.}
    \begin{tabular}{c|c|c}
    \toprule
    Hyperparameter  & Stage 1 & Stage 2 \\
    \midrule
    Learning rate   & 1e-5 & 1e-5 \\
    Warmup steps   & 800 & 400 \\
    LoRA Rank   & 64 & 64 \\
    No. of epochs   & 10 & 3\\
    Batch size   & 128 & 256 \\
    Number of samples per language & 20 hours & 90 000 \\
    \bottomrule
    \end{tabular}
    \label{tab:hyperparams}
\end{table}

In total, we train four models: (1) ALL+AST: a model trained on all of Aya's 23 languages with CoVoST-2 AST data; (2) ALL: a model trained on all of Aya's 23 languages; (3) TEN: a model trained with ten selected languages (English, French, Dutch, Turkish, German, Arabic, Spanish, Russian, Indonesian, and Polish); and (4) TEN+AST: a model trained with ten selected languages with CoVoST-2 AST data.

\begin{figure}
    \centering
    \includegraphics[width=\linewidth]{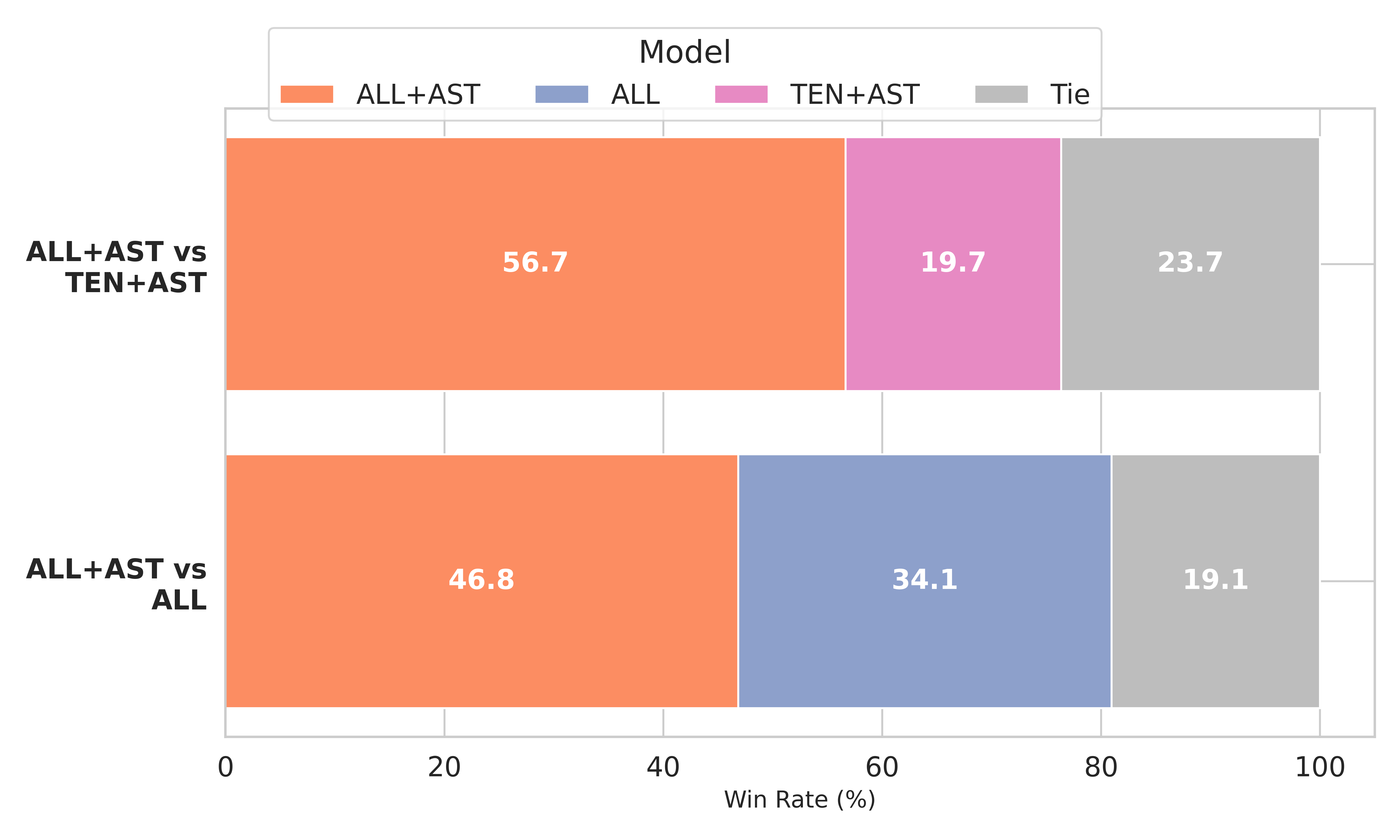}
    \caption{Win rates on \eval{} averaged across languages for the from-scratch SALMONN models. ALL+AST versus TEN+AST is shown at the top and versus ALL at the bottom. Bars show \% wins for each model and \% ties (gray).}
    \label{fig:number-of-langs}
\end{figure}

\subsection{Results}
\subsubsection{Does training with fewer languages lead to better performance?} Figure~\ref{fig:number-of-langs} summarises the difference in SQA performance of the models trained with 10 languages (TEN/TEN+AST) and 23 languages (ALL/ALL+AST). We see that the model trained with 23 languages results in a better win rate overall, suggesting that we do not experience capacity dilution at 23 languages, and adding more languages in training leads to better performance across languages. This could be due to the fact that we start with a pretrained speech encoder and a pretrained language model, meaning that we already have the question-answering capabilities present in the LM and the SLM training is primarily learning how to project the speech encoder output into the LM space.

\subsubsection{Does adding AST data improve the models?} We evaluate whether including speech translation data improves performance by testing on CoVoST-2 languages (both X to En and En to X translation directions) that overlap with the 23 languages in our dataset. We find that models trained with AST data win 46.8\% of the time against those without, indicating that additional AST data alone does not lead to a consistent improvement. This result is likely due to two factors: (1) the AST data comprises 20\% of the training mixture, which is potentially too small to produce a measurable effect; and (2) Whisper already supports speech translation into English, so adding a small amount of AST instruction-tuning data provides only limited additional supervision.

\section{Conclusion}

In this paper, we address the lack of multilingual instruction-tuning data for SLMs by presenting \dataset{}, a synthetic, human-verified dataset of more than 10.8 million instructions and 9200 hours of spoken question-answering data in 23 languages. We also introduce \eval{}, a human-verified, multilingual and multitask benchmark to evaluate SLMs on SQA, ASR and AST.
Using \eval{}, we establish the strong performance of Qwen2.5-Omni among the open-weight models we evaluate, and demonstrate the effectiveness of finetuning this model on \dataset{}, leading to state-of-the-art performance on the SQA portion of \eval{}.
These findings validate that automated, synthetic pipelines provide sufficient instruction-tuning data for effective post-training of SLMs across many languages.

\section{Limitations}

We present a synthetically generated dataset, which for several languages suggests the possibility of errors in the machine translation, which could lead to unnatural or possibly incorrect question prompts in our dataset. The quality of the generated speech is at the limit of the speech synthesis models, so for some languages, speech may be less natural. However, we ensure that the content can be understood by native speakers.

\section{Generative AI Use Disclosure}
Generative AI was used to generate plots and reformat tables.

\section{Acknowledgments}
This work is supported by the Stanford Interdisciplinary Graduate Fellowship, Google through the Stanford Institute for Human-Centered Artificial Intelligence (HAI). We thank Karen Livescu and Jordan Troutman, for their valuable comments and feedback, which have greatly strengthened this work.

Finally, we sincerely thank Cohere Labs community members participating our listening test:

Abdelaziz Bounhar, Aditya Punia, Ahmad, Aleksandra, Alper Balbay, Anjman Sikarwar, Antonia Karamolegkou, Aria Vaghayenegar, Arno Bourgonje, Arshia Soltani Moakhar, Ashay Srivastava, Carlos Miguel Patiño, Chen Shani, Claudia Quant, Constantinos Karouzos, Dante Lok, Danylo Boiko, Dipika Khullar, Dominic Liu, Dominik, Drishti Sharma, Efstathios Siatras, Eros Melo Barros, Esra'a Saleh, Fabian Farestam, Francisco Emiliano Lopez Saavedra, Gabriel da Costa Merlin, Gabriele Sarti, Gagandeep Kaur, Grace Chong, Guillaume de Malézieux, Haitame Laframe, Hanna Yukhymenko, Harsha, Hassan Alshanqiti, Heleen Prins, Hina, Jaeyoon Jung, Jafar Isbarov, Jebish Purbey, Jeron Bartelds, Joana da Matta, Jonibek Mansurov, Joseph Pollack, Julia Kreutzer, Jun Park, Kamshat Saduakassova, Kareem Ahmad, Kruthi.R.Vasishtha, Lauren Altomare, M, Malvina Nikandrou, Manuel Goulão, Marek Suppa, Michele Lugano, Micol Altomare, Mike Zhang, Morteza Kashani, Muhammad Hazim Al Farouq, Naman Bhatia, Nurdaulet Mukhituly, Otávio Ferracioli Coletti, Piyush, Prahitha Movva, Qianghao Wu, Rahul Dutta, Ram Mohan Rao Kadiyala, Reuben, Richard Nguyen, Rose3, Salsabila Zahirah, Sara Papi, Sarah Lintang, Sarthak Rawat, Sarvesh Gharat, Shafagh Fadaei, Shivam Garg, Shivansh Pachnanda, Siddharth Rawat, Silvia Fernandez, Simon de Wit, Vivek, Vlad Vasilescu, Yash Thube, Yashika Jain, Yiyang Nan, Yochai Shavit, Yunus Serhat Bicakci, Zahra Alharz.

\clearpage

\bibliographystyle{IEEEtran}
\bibliography{mybib}

@IEEEtranBSTCTL{IEEEexample:BSTcontrol,
  CTLuse_forced_etal = "yes",
  CTLmax_names_forced_etal = "10",
  CTLnames_show_etal = "10"
}

@misc{Jo2022ZerothKorean,
  author       = {Lucas Jo and Wonkyum Lee},
  title        = {Zeroth-Korean: Korean Open-source Speech Corpus for Speech Recognition},
  year         = {2022},
  url          = {https://openslr.org/40/},
  note         = {Accessed: 2025-09-21},
}

@article{diwan2021multilingual,
title={Multilingual and code-switching ASR challenges for low resource Indian languages},
author={Anuj Diwan and Rakesh Vaideeswaran and Sanket Shah and Ankita Singh and Srinivasa Raghavan and Shreya Khare and Vinit Unni and Saurabh Vyas and Akash Rajpuria and Chiranjeevi Yarra and Ashish Mittal and Prasanta Kumar Ghosh and Preethi Jyothi and Kalika Bali and Vivek Seshadri and Sunayana Sitaram and Samarth Bharadwaj and Jai Nanavati and Raoul Nanavati and Karthik Sankaranarayanan and Tejaswi Seeram and Basil Abraham},
Journal={Proceedings of Interspeech},
year={2021}
}

@article{chu2024qwen2,
  title={Qwen2-audio technical report},
  author={Chu, Yunfei and Xu, Jin and Yang, Qian and Wei, Haojie and Wei, Xipin and Guo, Zhifang and Leng, Yichong and Lv, Yuanjun and He, Jinzheng and Lin, Junyang and others},
  journal={arXiv preprint arXiv:2407.10759},
  year={2024}
}

@misc{dang2024ayaexpansecombiningresearch,
      title={Aya Expanse: Combining Research Breakthroughs for a New Multilingual Frontier}, 
      author={John Dang and Shivalika Singh and Daniel D'souza and Arash Ahmadian and Alejandro Salamanca and Madeline Smith and Aidan Peppin and Sungjin Hong and Manoj Govindassamy and Terrence Zhao and Sandra Kublik and Meor Amer and Viraat Aryabumi and Jon Ander Campos and Yi-Chern Tan and Tom Kocmi and Florian Strub and Nathan Grinsztajn and Yannis Flet-Berliac and Acyr Locatelli and Hangyu Lin and Dwarak Talupuru and Bharat Venkitesh and David Cairuz and Bowen Yang and Tim Chung and Wei-Yin Ko and Sylvie Shang Shi and Amir Shukayev and Sammie Bae and Aleksandra Piktus and Roman Castagné and Felipe Cruz-Salinas and Eddie Kim and Lucas Crawhall-Stein and Adrien Morisot and Sudip Roy and Phil Blunsom and Ivan Zhang and Aidan Gomez and Nick Frosst and Marzieh Fadaee and Beyza Ermis and Ahmet Üstün and Sara Hooker},
      year={2024},
      eprint={2412.04261},
      archivePrefix={arXiv},
      primaryClass={cs.CL},
      url={https://arxiv.org/abs/2412.04261}, 
}

@article{cohere2025command,
  title={Command {A}: An enterprise-ready large language model},
  author={Cohere, Team and Ahmadian, Arash and Ahmed, Marwan and Alammar, Jay and Alizadeh, Milad and Alnumay, Yazeed and Althammer, Sophia and Arkhangorodsky, Arkady and Aryabumi, Viraat and Aumiller, Dennis and others},
  journal={arXiv preprint arXiv:2504.00698},
  year={2025}
}

@article{peng2023instruction,
  title={Instruction Tuning with GPT-4},
  author={Peng, Baolin and Li, Chunyuan and He, Pengcheng and Galley, Michel and Gao, Jianfeng},
  journal={arXiv preprint arXiv:2304.03277},
  year={2023}
}

@article{hh-rlhf-anthropic,
  title={Training a Helpful and Harmless Assistant with Reinforcement Learning from Human Feedback},
  author={Bai, Yuntao and Jones, Andy and Ndousse, Kamal and Askell, Amanda and Chen, Anna and DasSarma, Nova and Drain, Dawn and Fort, Stanislav and Ganguli, Deep and Henighan, Tom and others},
  journal={CoRR},
  year={2022}
}

@misc {trvia-single-choice,
  author = {Mihai},
  title = {trivia-single-choice},
  url = {https://huggingface.co/datasets/Mihaiii/trivia_single_choice},
  type = {dataset},
  year = {2024},
  month = {July},
  note = {Accessed: 2025-09-17}
}

@misc {qa-assitant,
  author = {Mihai},
  title = {qa-assistant},
  url = {https://huggingface.co/datasets/Mihaiii/qa-assistant},
  type = {dataset},
  year = {2024},
  month = {April},
  note = {Accessed: 2025-09-17}
}

@misc {qa-assitant-2,
  author = {Mihai},
  title = {qa-assistant-2},
  url = {https://huggingface.co/datasets/Mihaiii/qa-assistant-2},
  type = {dataset},
  year = {2024},
  month = {June},
  note = {Accessed: 2025-09-17}
}

@article{xie2024mini,
  title={Mini-omni: Language models can hear, talk while thinking in streaming},
  author={Xie, Zhifei and Wu, Changqiao},
  journal={arXiv preprint arXiv:2408.16725},
  year={2024}
}

@inproceedings{seamless2023,
   title = {Seamless Communication},
  author = "{Seamless Communication} and Lo{\"i}c Barrault and Yu-An Chung and Mariano Coria Meglioli and David Dale and Ning Dong and Mark Duppenthaler and Paul-Ambroise Duquenne and Brian Ellis and Hady Elsahar and Justin Haaheim and John Hoffman and Min-Jae Hwang and Hirofumi Inaguma and Christopher Klaiber and Ilia Kulikov and Pengwei Li and Daniel Licht and Jean Maillard and Ruslan Mavlyutov and Alice Rakotoarison and Kaushik Ram Sadagopan and Abinesh Ramakrishnan and Tuan Tran and Guillaume Wenzek and Yilin Yang and Ethan Ye and Ivan Evtimov and Pierre Fernandez and Cynthia Gao and Prangthip Hansanti and Elahe Kalbassi and Amanda Kallet and Artyom Kozhevnikov and Gabriel Mejia and Robin San Roman and Christophe Touret and Corinne Wong and Carleigh Wood and Bokai Yu and Pierre Andrews and Can Balioglu and Peng-Jen Chen and Marta R. Costa-juss{\`a} and Maha Elbayad and Hongyu Gong and Francisco Guzm{\'a}n and Kevin Heffernan and Somya Jain and Justine Kao and Ann Lee and Xutai Ma and Alex Mourachko and Benjamin Peloquin and Juan Pino and Sravya Popuri and Christophe Ropers and Safiyyah Saleem and Holger Schwenk and Anna Sun and Paden Tomasello and Changhan Wang and Jeff Wang and Skyler Wang and Mary Williamson",
  journal={ArXiv},
  year={2023}
}

@inproceedings{casanova2024xtts,
  title={XTTS: a Massively Multilingual Zero-Shot Text-to-Speech Model},
  author={Casanova, Edresson and Davis, Kelly and G{\"o}lge, Eren and G{\"o}knar, G{\"o}rkem and Gulea, Iulian and Hart, Logan and Aljafari, Aya and Meyer, Joshua and Morais, Reuben and Olayemi, Samuel and others},
  booktitle={Proc. Interspeech 2024},
  pages={4978--4982},
  year={2024}
}

@article{pratap2024scaling,
  title={Scaling speech technology to 1,000+ languages},
  author={Pratap, Vineel and Tjandra, Andros and Shi, Bowen and Tomasello, Paden and Babu, Arun and Kundu, Sayani and Elkahky, Ali and Ni, Zhaoheng and Vyas, Apoorv and Fazel-Zarandi, Maryam and others},
  journal={Journal of Machine Learning Research},
  volume={25},
  number={97},
  pages={1--52},
  year={2024}
}

@inproceedings{radford2023robust,
  title={Robust speech recognition via large-scale weak supervision},
  author={Radford, Alec and Kim, Jong Wook and Xu, Tao and Brockman, Greg and McLeavey, Christine and Sutskever, Ilya},
  booktitle={International conference on machine learning},
  pages={28492--28518},
  year={2023},
  organization={PMLR}
}

@article{abouelenin2025phi,
  title={Phi-4-mini technical report: Compact yet powerful multimodal language models via mixture-of-loras},
  author={Abouelenin, Abdelrahman and Ashfaq, Atabak and Atkinson, Adam and Awadalla, Hany and Bach, Nguyen and Bao, Jianmin and Benhaim, Alon and Cai, Martin and Chaudhary, Vishrav and Chen, Congcong and others},
  journal={arXiv preprint arXiv:2503.01743},
  year={2025}
}

@misc{arora2025landscapespokenlanguagemodels,
      title={{On The Landscape of Spoken Language Models: A Comprehensive Survey}}, 
      author={Siddhant Arora and Kai-Wei Chang and Chung-Ming Chien and Yifan Peng and Haibin Wu and Yossi Adi and Emmanuel Dupoux and Hung-Yi Lee and Karen Livescu and Shinji Watanabe},
      year={2025},
      eprint={2504.08528},
      archivePrefix={arXiv},
      primaryClass={cs.CL},
      url={https://arxiv.org/abs/2504.08528}, 
}

@inproceedings{shi24g_interspeech,
  title     = {ML-SUPERB 2.0: Benchmarking Multilingual Speech Models Across Modeling Constraints, Languages, and Datasets},
  author    = {Jiatong Shi and Shih-Heng Wang and William Chen and Martijn Bartelds and Vanya {Bannihatti Kumar} and Jinchuan Tian and Xuankai Chang and Dan Jurafsky and Karen Livescu and Hung-yi Lee and Shinji Watanabe},
  year      = {2024},
  booktitle = {Interspeech 2024},
  pages     = {1230--1234},
  doi       = {10.21437/Interspeech.2024-2248},
  issn      = {2958-1796},
}

@misc{openai2024gpt4technicalreport,
      title={{GPT-4 Technical Report}}, 
      author={OpenAI and Josh Achiam and Steven Adler and Sandhini Agarwal and Lama Ahmad and Ilge Akkaya and Florencia Leoni Aleman and Diogo Almeida and Janko Altenschmidt and Sam Altman and Shyamal Anadkat and Red Avila and Igor Babuschkin and Suchir Balaji and Valerie Balcom and Paul Baltescu and Haiming Bao and Mohammad Bavarian and Jeff Belgum and Irwan Bello and Jake Berdine and Gabriel Bernadett-Shapiro and Christopher Berner and Lenny Bogdonoff and Oleg Boiko and Madelaine Boyd and Anna-Luisa Brakman and Greg Brockman and Tim Brooks and Miles Brundage and Kevin Button and Trevor Cai and Rosie Campbell and Andrew Cann and Brittany Carey and Chelsea Carlson and Rory Carmichael and Brooke Chan and Che Chang and Fotis Chantzis and Derek Chen and Sully Chen and Ruby Chen and Jason Chen and Mark Chen and Ben Chess and Chester Cho and Casey Chu and Hyung Won Chung and Dave Cummings and Jeremiah Currier and Yunxing Dai and Cory Decareaux and Thomas Degry and Noah Deutsch and Damien Deville and Arka Dhar and David Dohan and Steve Dowling and Sheila Dunning and Adrien Ecoffet and Atty Eleti and Tyna Eloundou and David Farhi and Liam Fedus and Niko Felix and Simón Posada Fishman and Juston Forte and Isabella Fulford and Leo Gao and Elie Georges and Christian Gibson and Vik Goel and Tarun Gogineni and Gabriel Goh and Rapha Gontijo-Lopes and Jonathan Gordon and Morgan Grafstein and Scott Gray and Ryan Greene and Joshua Gross and Shixiang Shane Gu and Yufei Guo and Chris Hallacy and Jesse Han and Jeff Harris and Yuchen He and Mike Heaton and Johannes Heidecke and Chris Hesse and Alan Hickey and Wade Hickey and Peter Hoeschele and Brandon Houghton and Kenny Hsu and Shengli Hu and Xin Hu and Joost Huizinga and Shantanu Jain and Shawn Jain and Joanne Jang and Angela Jiang and Roger Jiang and Haozhun Jin and Denny Jin and Shino Jomoto and Billie Jonn and Heewoo Jun and Tomer Kaftan and Łukasz Kaiser and Ali Kamali and Ingmar Kanitscheider and Nitish Shirish Keskar and Tabarak Khan and Logan Kilpatrick and Jong Wook Kim and Christina Kim and Yongjik Kim and Jan Hendrik Kirchner and Jamie Kiros and Matt Knight and Daniel Kokotajlo and Łukasz Kondraciuk and Andrew Kondrich and Aris Konstantinidis and Kyle Kosic and Gretchen Krueger and Vishal Kuo and Michael Lampe and Ikai Lan and Teddy Lee and Jan Leike and Jade Leung and Daniel Levy and Chak Ming Li and Rachel Lim and Molly Lin and Stephanie Lin and Mateusz Litwin and Theresa Lopez and Ryan Lowe and Patricia Lue and Anna Makanju and Kim Malfacini and Sam Manning and Todor Markov and Yaniv Markovski and Bianca Martin and Katie Mayer and Andrew Mayne and Bob McGrew and Scott Mayer McKinney and Christine McLeavey and Paul McMillan and Jake McNeil and David Medina and Aalok Mehta and Jacob Menick and Luke Metz and Andrey Mishchenko and Pamela Mishkin and Vinnie Monaco and Evan Morikawa and Daniel Mossing and Tong Mu and Mira Murati and Oleg Murk and David Mély and Ashvin Nair and Reiichiro Nakano and Rajeev Nayak and Arvind Neelakantan and Richard Ngo and Hyeonwoo Noh and Long Ouyang and Cullen O'Keefe and Jakub Pachocki and Alex Paino and Joe Palermo and Ashley Pantuliano and Giambattista Parascandolo and Joel Parish and Emy Parparita and Alex Passos and Mikhail Pavlov and Andrew Peng and Adam Perelman and Filipe de Avila Belbute Peres and Michael Petrov and Henrique Ponde de Oliveira Pinto and Michael and Pokorny and Michelle Pokrass and Vitchyr H. Pong and Tolly Powell and Alethea Power and Boris Power and Elizabeth Proehl and Raul Puri and Alec Radford and Jack Rae and Aditya Ramesh and Cameron Raymond and Francis Real and Kendra Rimbach and Carl Ross and Bob Rotsted and Henri Roussez and Nick Ryder and Mario Saltarelli and Ted Sanders and Shibani Santurkar and Girish Sastry and Heather Schmidt and David Schnurr and John Schulman and Daniel Selsam and Kyla Sheppard and Toki Sherbakov and Jessica Shieh and Sarah Shoker and Pranav Shyam and Szymon Sidor and Eric Sigler and Maddie Simens and Jordan Sitkin and Katarina Slama and Ian Sohl and Benjamin Sokolowsky and Yang Song and Natalie Staudacher and Felipe Petroski Such and Natalie Summers and Ilya Sutskever and Jie Tang and Nikolas Tezak and Madeleine B. Thompson and Phil Tillet and Amin Tootoonchian and Elizabeth Tseng and Preston Tuggle and Nick Turley and Jerry Tworek and Juan Felipe Cerón Uribe and Andrea Vallone and Arun Vijayvergiya and Chelsea Voss and Carroll Wainwright and Justin Jay Wang and Alvin Wang and Ben Wang and Jonathan Ward and Jason Wei and CJ Weinmann and Akila Welihinda and Peter Welinder and Jiayi Weng and Lilian Weng and Matt Wiethoff and Dave Willner and Clemens Winter and Samuel Wolrich and Hannah Wong and Lauren Workman and Sherwin Wu and Jeff Wu and Michael Wu and Kai Xiao and Tao Xu and Sarah Yoo and Kevin Yu and Qiming Yuan and Wojciech Zaremba and Rowan Zellers and Chong Zhang and Marvin Zhang and Shengjia Zhao and Tianhao Zheng and Juntang Zhuang and William Zhuk and Barret Zoph},
      year={2024},
      eprint={2303.08774},
      archivePrefix={arXiv},
      primaryClass={cs.CL},
      url={https://arxiv.org/abs/2303.08774}, 
}

@INPROCEEDINGS{fleurs,
  author={Conneau, Alexis and Ma, Min and Khanuja, Simran and Zhang, Yu and Axelrod, Vera and Dalmia, Siddharth and Riesa, Jason and Rivera, Clara and Bapna, Ankur},
  booktitle={2022 IEEE Spoken Language Technology Workshop (SLT)}, 
  title={FLEURS: FEW-Shot Learning Evaluation of Universal Representations of Speech}, 
  year={2023},
  volume={},
  number={},
  pages={798-805},
  doi={10.1109/SLT54892.2023.10023141}}

@article{Qwen2-Audio,
  title={Qwen2-Audio Technical Report},
  author={Chu, Yunfei and Xu, Jin and Yang, Qian and Wei, Haojie and Wei, Xipin and Guo,  Zhifang and Leng, Yichong and Lv, Yuanjun and He, Jinzheng and Lin, Junyang and Zhou, Chang and Zhou, Jingren},
  journal={arXiv preprint arXiv:2407.10759},
  year={2024}
}

@inproceedings{
  salmonn,
  title={{SALMONN}: Towards Generic Hearing Abilities for Large Language Models},
  author={Changli Tang and Wenyi Yu and Guangzhi Sun and Xianzhao Chen and Tian Tan and Wei Li and Lu Lu and Zejun MA and Chao Zhang},
  booktitle={The Twelfth International Conference on Learning Representations},
  year={2024},
  url={https://openreview.net/forum?id=14rn7HpKVk}
}

@inproceedings{commonvoice,
    title = "Common Voice: A Massively-Multilingual Speech Corpus",
    author = "Ardila, Rosana  and
      Branson, Megan  and
      Davis, Kelly  and
      Kohler, Michael  and
      Meyer, Josh  and
      Henretty, Michael  and
      Morais, Reuben  and
      Saunders, Lindsay  and
      Tyers, Francis  and
      Weber, Gregor",
    editor = "Calzolari, Nicoletta  and
      B{\'e}chet, Fr{\'e}d{\'e}ric  and
      Blache, Philippe  and
      Choukri, Khalid  and
      Cieri, Christopher  and
      Declerck, Thierry  and
      Goggi, Sara  and
      Isahara, Hitoshi  and
      Maegaard, Bente  and
      Mariani, Joseph  and
      Mazo, H{\'e}l{\`e}ne  and
      Moreno, Asuncion  and
      Odijk, Jan  and
      Piperidis, Stelios",
    booktitle = "Proceedings of the Twelfth Language Resources and Evaluation Conference",
    month = may,
    year = "2020",
    address = "Marseille, France",
    publisher = "European Language Resources Association",
    url = "https://aclanthology.org/2020.lrec-1.520/",
    pages = "4218--4222",
    language = "eng",
    ISBN = "979-10-95546-34-4"
}

@misc{Bud500,
  author = {Anh Pham and Khanh Linh Tran and Linh Nguyen and Thanh Duy Cao and Phuc Phan and Duong A. Nguyen},
  title = {Bud500: A Comprehensive Vietnamese ASR Dataset},
  url = {https://github.com/quocanh34/Bud500},
  year = {2024}
}

@misc{marmor2023ivritai,
      title={ivrit.ai: A Comprehensive Dataset of Hebrew Speech for AI Research and Development}, 
      author={Yanir Marmor and Kinneret Misgav and Yair Lifshitz},
      year={2023},
      eprint={2307.08720},
      archivePrefix={arXiv},
      primaryClass={eess.AS}
}

@inproceedings{zhang2023speechgpt,
  title={SpeechGPT: Empowering Large Language Models with Intrinsic Cross-Modal Conversational Abilities},
  author={Zhang, Dong and Li, Shimin and Zhang, Xin and Zhan, Jun and Wang, Pengyu and Zhou, Yaqian and Qiu, Xipeng},
  booktitle={Findings of the Association for Computational Linguistics: EMNLP 2023},
  pages={15757--15773},
  year={2023}
}

@article{fang2024llama,
  title={LLaMA-Omni: Seamless Speech Interaction with Large Language Models},
  author={Fang, Qingkai and Guo, Shoutao and Zhou, Yan and Ma, Zhengrui and Zhang, Shaolei and Feng, Yang},
  journal={CoRR},
  year={2024}
}

@inproceedings{yue2025pangea,
title={Pangea: A Fully Open Multilingual Multimodal {LLM} for 39 Languages},
author={Xiang Yue and Yueqi Song and Akari Asai and Seungone Kim and Jean de Dieu Nyandwi and Simran Khanuja and Anjali Kantharuban and Lintang Sutawika and Sathyanarayanan Ramamoorthy and Graham Neubig},
booktitle={The Thirteenth International Conference on Learning Representations},
year={2025},
url={https://openreview.net/forum?id=a3g2l4yEys}
}

@inproceedings{singh-etal-2024-aya,
    title = "Aya Dataset: An Open-Access Collection for Multilingual Instruction Tuning",
    author = {Singh, Shivalika  and
      Vargus, Freddie  and
      D{'}souza, Daniel  and
      Karlsson, B{\"o}rje  and
      Mahendiran, Abinaya  and
      Ko, Wei-Yin  and
      Shandilya, Herumb  and
      Patel, Jay  and
      Mataciunas, Deividas  and
      O{'}Mahony, Laura  and
      Zhang, Mike  and
      Hettiarachchi, Ramith  and
      Wilson, Joseph  and
      Machado, Marina  and
      Moura, Luisa  and
      Krzemi{\'n}ski, Dominik  and
      Fadaei, Hakimeh  and
      Ergun, Irem  and
      Okoh, Ifeoma  and
      Alaagib, Aisha  and
      Mudannayake, Oshan  and
      Alyafeai, Zaid  and
      Chien, Vu  and
      Ruder, Sebastian  and
      Guthikonda, Surya  and
      Alghamdi, Emad  and
      Gehrmann, Sebastian  and
      Muennighoff, Niklas  and
      Bartolo, Max  and
      Kreutzer, Julia  and
      {\"U}st{\"u}n, Ahmet  and
      Fadaee, Marzieh  and
      Hooker, Sara},
    editor = "Ku, Lun-Wei  and
      Martins, Andre  and
      Srikumar, Vivek",
    booktitle = "Proceedings of the 62nd Annual Meeting of the Association for Computational Linguistics (Volume 1: Long Papers)",
    month = aug,
    year = "2024",
    address = "Bangkok, Thailand",
    publisher = "Association for Computational Linguistics",
    url = "https://aclanthology.org/2024.acl-long.620/",
    doi = "10.18653/v1/2024.acl-long.620",
    pages = "11521--11567"
}

@inproceedings{ustun-etal-2024-aya,
    title = "Aya Model: An Instruction Finetuned Open-Access Multilingual Language Model",
    author = {{\"U}st{\"u}n, Ahmet  and
      Aryabumi, Viraat  and
      Yong, Zheng  and
      Ko, Wei-Yin  and
      D{'}souza, Daniel  and
      Onilude, Gbemileke  and
      Bhandari, Neel  and
      Singh, Shivalika  and
      Ooi, Hui-Lee  and
      Kayid, Amr  and
      Vargus, Freddie  and
      Blunsom, Phil  and
      Longpre, Shayne  and
      Muennighoff, Niklas  and
      Fadaee, Marzieh  and
      Kreutzer, Julia  and
      Hooker, Sara},
    editor = "Ku, Lun-Wei  and
      Martins, Andre  and
      Srikumar, Vivek",
    booktitle = "Proceedings of the 62nd Annual Meeting of the Association for Computational Linguistics (Volume 1: Long Papers)",
    month = aug,
    year = "2024",
    address = "Bangkok, Thailand",
    publisher = "Association for Computational Linguistics",
    url = "https://aclanthology.org/2024.acl-long.845/",
    doi = "10.18653/v1/2024.acl-long.845",
    pages = "15894--15939"
}

@article{dash2025aya,
  title={Aya Vision: Advancing the Frontier of Multilingual Multimodality},
  author={Dash, Saurabh and Nan, Yiyang and Dang, John and Ahmadian, Arash and Singh, Shivalika and Smith, Madeline and Venkitesh, Bharat and Shmyhlo, Vlad and Aryabumi, Viraat and Beller-Morales, Walter and others},
  journal={arXiv preprint arXiv:2505.08751},
  year={2025}
}

@article{team2022no,
  title={No language left behind: Scaling human-centered machine translation},
  author={Team, NLLB and Costa-Juss{\`a}, Marta R and Cross, James and {\c{C}}elebi, Onur and Elbayad, Maha and Heafield, Kenneth and Heffernan, Kevin and Kalbassi, Elahe and Lam, Janice and Licht, Daniel and others},
  journal={arXiv preprint arXiv:2207.04672},
  year={2022},
  publisher={Aug}
}

@article{comanici2025gemini,
  title={Gemini 2.5: Pushing the frontier with advanced reasoning, multimodality, long context, and next generation agentic capabilities},
  author={Comanici, Gheorghe and Bieber, Eric and Schaekermann, Mike and Pasupat, Ice and Sachdeva, Noveen and Dhillon, Inderjit and Blistein, Marcel and Ram, Ori and Zhang, Dan and Rosen, Evan and others},
  journal={arXiv preprint arXiv:2507.06261},
  year={2025}
}

@inproceedings{wang2021covost,
  title={CoVoST 2 and massively multilingual speech translation.},
  author={Wang, Changhan and Wu, Anne and Gu, Jiatao and Pino, Juan},
  booktitle={Interspeech},
  volume={2021},
  pages={2247--2251},
  year={2021}
}

@article{Qwen2.5-Omni,
  title={Qwen2.5-Omni Technical Report},
  author={Jin Xu and Zhifang Guo and Jinzheng He and Hangrui Hu and Ting He and Shuai Bai and Keqin Chen and Jialin Wang and Yang Fan and Kai Dang and Bin Zhang and Xiong Wang and Yunfei Chu and Junyang Lin},
  journal={arXiv preprint arXiv:2503.20215},
  year={2025}
}

@misc{McGuire_2005, title={Librivox}, url={https://librivox.org/}, journal={LibriVox}, author={McGuire, Hugh}, year={2005}, month={Aug}}

@misc{chen2024voicebenchbenchmarkingllmbasedvoice,
      title={{VoiceBench: Benchmarking LLM-Based Voice Assistants}}, 
      author={Yiming Chen and Xianghu Yue and Chen Zhang and Xiaoxue Gao and Robby T. Tan and Haizhou Li},
      year={2024},
      eprint={2410.17196},
      archivePrefix={arXiv},
      primaryClass={cs.CL},
      url={https://arxiv.org/abs/2410.17196}, 
}

@inproceedings{NEURIPS2018_6832a7b2,
 author = {Jia, Ye and Zhang, Yu and Weiss, Ron and Wang, Quan and Shen, Jonathan and Ren, Fei and Chen, zhifeng and Nguyen, Patrick and Pang, Ruoming and Lopez Moreno, Ignacio and Wu, Yonghui},
 booktitle = {Advances in Neural Information Processing Systems},
 editor = {S. Bengio and H. Wallach and H. Larochelle and K. Grauman and N. Cesa-Bianchi and R. Garnett},
 pages = {},
 publisher = {Curran Associates, Inc.},
 title = {Transfer Learning from Speaker Verification to Multispeaker Text-To-Speech Synthesis},
 url = {https://proceedings.neurips.cc/paper_files/paper/2018/file/6832a7b24bc06775d02b7406880b93fc-Paper.pdf},
 volume = {31},
 year = {2018}
}

@inproceedings{hu2022lora,
title={Lo{RA}: Low-Rank Adaptation of Large Language Models},
author={Edward J Hu and yelong shen and Phillip Wallis and Zeyuan Allen-Zhu and Yuanzhi Li and Shean Wang and Lu Wang and Weizhu Chen},
booktitle={International Conference on Learning Representations},
year={2022},
url={https://openreview.net/forum?id=nZeVKeeFYf9}
}

@inproceedings{papineni-etal-2002-bleu,
    title = "{B}leu: a Method for Automatic Evaluation of Machine Translation",
    author = "Papineni, Kishore  and
      Roukos, Salim  and
      Ward, Todd  and
      Zhu, Wei-Jing",
    editor = "Isabelle, Pierre  and
      Charniak, Eugene  and
      Lin, Dekang",
    booktitle = "Proceedings of the 40th Annual Meeting of the Association for Computational Linguistics",
    month = jul,
    year = "2002",
    address = "Philadelphia, Pennsylvania, USA",
    publisher = "Association for Computational Linguistics",
    url = "https://aclanthology.org/P02-1040/",
    doi = "10.3115/1073083.1073135",
    pages = "311--318"
}

@inproceedings{popovic-2015-chrf,
    title = "chr{F}: character n-gram {F}-score for automatic {MT} evaluation",
    author = "Popovi{\'c}, Maja",
    editor = "Bojar, Ond{\v{r}}ej  and
      Chatterjee, Rajan  and
      Federmann, Christian  and
      Haddow, Barry  and
      Hokamp, Chris  and
      Huck, Matthias  and
      Logacheva, Varvara  and
      Pecina, Pavel",
    booktitle = "Proceedings of the Tenth Workshop on Statistical Machine Translation",
    month = sep,
    year = "2015",
    address = "Lisbon, Portugal",
    publisher = "Association for Computational Linguistics",
    url = "https://aclanthology.org/W15-3049/",
    doi = "10.18653/v1/W15-3049",
    pages = "392--395"
}

@article{wang2024audiobench,
    title={AudioBench: A Universal Benchmark for Audio Large Language Models},
    author={Wang, Bin and Zou, Xunlong and Lin, Geyu and Sun, Shuo and Liu, Zhuohan and Zhang, Wenyu and Liu, Zhengyuan and Aw, AiTi and Chen, Nancy F},
    journal={NAACL},
    year={2025}
    }

@inproceedings{shon2022slue,
  title={Slue: New benchmark tasks for spoken language understanding evaluation on natural speech},
  author={Shon, Suwon and Pasad, Ankita and Wu, Felix and Brusco, Pablo and Artzi, Yoav and Livescu, Karen and Han, Kyu J},
  booktitle={ICASSP 2022-2022 IEEE International Conference on Acoustics, Speech and Signal Processing (ICASSP)},
  pages={7927--7931},
  year={2022},
  organization={IEEE}
}

@article{lee2025ahelm,
  title={Ahelm: A holistic evaluation of audio-language models},
  author={Lee, Tony and Tu, Haoqin and Wong, Chi Heem and Wang, Zijun and Yang, Siwei and Mai, Yifan and Zhou, Yuyin and Xie, Cihang and Liang, Percy},
  journal={arXiv preprint arXiv:2508.21376},
  year={2025}
}

@article{yang2024air,
  title={Air-bench: Benchmarking large audio-language models via generative comprehension},
  author={Yang, Qian and Xu, Jin and Liu, Wenrui and Chu, Yunfei and Jiang, Ziyue and Zhou, Xiaohuan and Leng, Yichong and Lv, Yuanjun and Zhao, Zhou and Zhou, Chang and others},
  journal={arXiv preprint arXiv:2402.07729},
  year={2024}
}

@misc{ahia2025blabbrutallylongaudio,
      title={{BLAB: Brutally Long Audio Bench}}, 
      author={Orevaoghene Ahia and Martijn Bartelds and Kabir Ahuja and Hila Gonen and Valentin Hofmann and Siddhant Arora and Shuyue Stella Li and Vishal Puttagunta and Mofetoluwa Adeyemi and Charishma Buchireddy and Ben Walls and Noah Bennett and Shinji Watanabe and Noah A. Smith and Yulia Tsvetkov and Sachin Kumar},
      year={2025},
      eprint={2505.03054},
      archivePrefix={arXiv},
      primaryClass={cs.AI},
      url={https://arxiv.org/abs/2505.03054}, 
}

\end{document}